\documentclass{article}

    \PassOptionsToPackage{numbers, compress}{natbib}

    \usepackage[final]{neurips_2024}

\usepackage[utf8]{inputenc} 
\usepackage[T1]{fontenc}    
\usepackage{hyperref}       
\usepackage{url}            
\usepackage{booktabs}       
\usepackage{amsfonts}       
\usepackage{nicefrac}       
\usepackage{microtype}      
\usepackage{xcolor}         

\usepackage{algorithmic}
\usepackage{algorithm}
\usepackage{amsfonts}
\usepackage{amsmath}
\usepackage{amssymb}
\usepackage{amsthm,wrapfig}
\usepackage{array}
\usepackage{arydshln}
\usepackage{bm}
\usepackage{cite}
\usepackage{color}
\usepackage{comment}
\usepackage{dsfont}
\usepackage{enumitem}
\usepackage{float}
\usepackage[T1]{fontenc}
\usepackage{geometry}
\usepackage{graphicx}
\usepackage{grffile}
\usepackage{hyperref}\hypersetup{colorlinks,linkcolor=blue,anchorcolor=blue,citecolor=blue}
\usepackage{letltxmacro}
\usepackage{mathrsfs}
\usepackage{mathtools}
\usepackage{multirow}
\usepackage{nicefrac}
\usepackage{subcaption}
\usepackage{tablefootnote}
\usepackage{verbatim}

\usepackage{todonotes}

\usepackage{booktabs}

\allowdisplaybreaks

\newcommand{\bb}{\bm{b}}

\newcommand{\bbm}{\bm{m}}

\newcommand{\bx}{\bm{x}}

\newcommand{\bM}{\bm{M}}

\newcommand{\bR}{\bm{R}}

\newcommand{\bW}{\bm{W}}
\newcommand{\bX}{\bm{X}}
\newcommand{\bY}{\bm{Y}}

\newcommand{\cC}{\mathcal{C}}
\newcommand{\cD}{\mathcal{D}}

\newcommand{\RR}{\mathbb{R}}

\title{Towards Low-bit Communication for\\Tensor Parallel LLM Inference}

\author{%
  Harry Dong\thanks{Work done at Apple.} \\
  Carnegie Mellon University \\
  \texttt{harryd@andrew.cmu.edu} \\
  \And
  Tyler Johnson \\
  Apple \\
  \And
  Minsik Cho \\
  Apple \\
  \And
  Emad Soroush \\
  Apple \\
}

\begin{document}

\maketitle

\begin{abstract}
Tensor parallelism provides an effective way to increase server large language model (LLM) inference efficiency despite adding an additional communication cost. However, as server LLMs continue to scale in size, they will need to be distributed across more devices, magnifying the communication cost.
One way to approach this problem is with quantization, but current methods for LLMs tend to avoid quantizing the features that tensor parallelism needs to communicate. Taking advantage of consistent outliers in communicated features, we introduce a quantization method that reduces communicated values on average from 16 bits to 4.2 bits while preserving nearly all of the original performance. For instance, our method maintains around 98.0\% and 99.5\% of Gemma 2 27B’s and Llama 2 13B's original performance, respectively, averaged across all tasks we evaluated on.
\end{abstract}

\section{Introduction}
\label{sec:introduction}

The use of large language models (LLMs) \citep{dubey2024llama,gunter2024apple,team2024gemma,mistralnemo} has ballooned in countless areas due to their impressive capabilities. Even so, with the enormous size of server LLMs, inference-time efficiency becomes a dire issue for those who own the models and those who use them. Fortunately, techniques like sequence parallelism \citep{li2021sequence} and tensor parallelism \citep{narayanan2021efficient, shoeybi2019megatron} distribute the computational load of transformer-based LLMs \citep{vaswani2017attention} onto different devices. However, these methods require communication between devices, which is especially a concern when serving models, so cheaper networking would greatly cut costs. In addition, as LLMs increase in size, we need to distribute them over more devices, which further drives up communication costs.

A natural idea is to try quantization, but this comes with challenges. Quantization methods for LLMs have largely focused on weights \citep{dettmers2023case, frantar2022gptq, lin2024awq, park2022lut, sheng2023flexgen} or multiplication between low-bit weights and low-bit input features \citep{ashkboos2024quarot, dettmers2022gpt3, wei2022outlier, xiao2023smoothquant, zhang2024lqer}. These methods are most useful for hardware-constrained settings, but their savings are not as relevant to tensor parallel server LLMs. In particular, current LLM quantization methods keep the output features of each attention and feedforward block at high precision (BF16 or FP16) to preserve performance. However, this is exactly what needs to be communicated in tensor parallelism. There has been work in quantized communication for distributed inference \citep{mitra2021distributed, nadendla2014distributed, qin2024disco}, but applications in LLMs have been limited. \textit{Consequently, the main challenge with quantization is to find a way to communicate low-bit output features from tensor parallelized attention and feedforward blocks across devices while preserving the original model's performance.}

\begin{figure}
\centering
\includegraphics[width=0.38\linewidth]{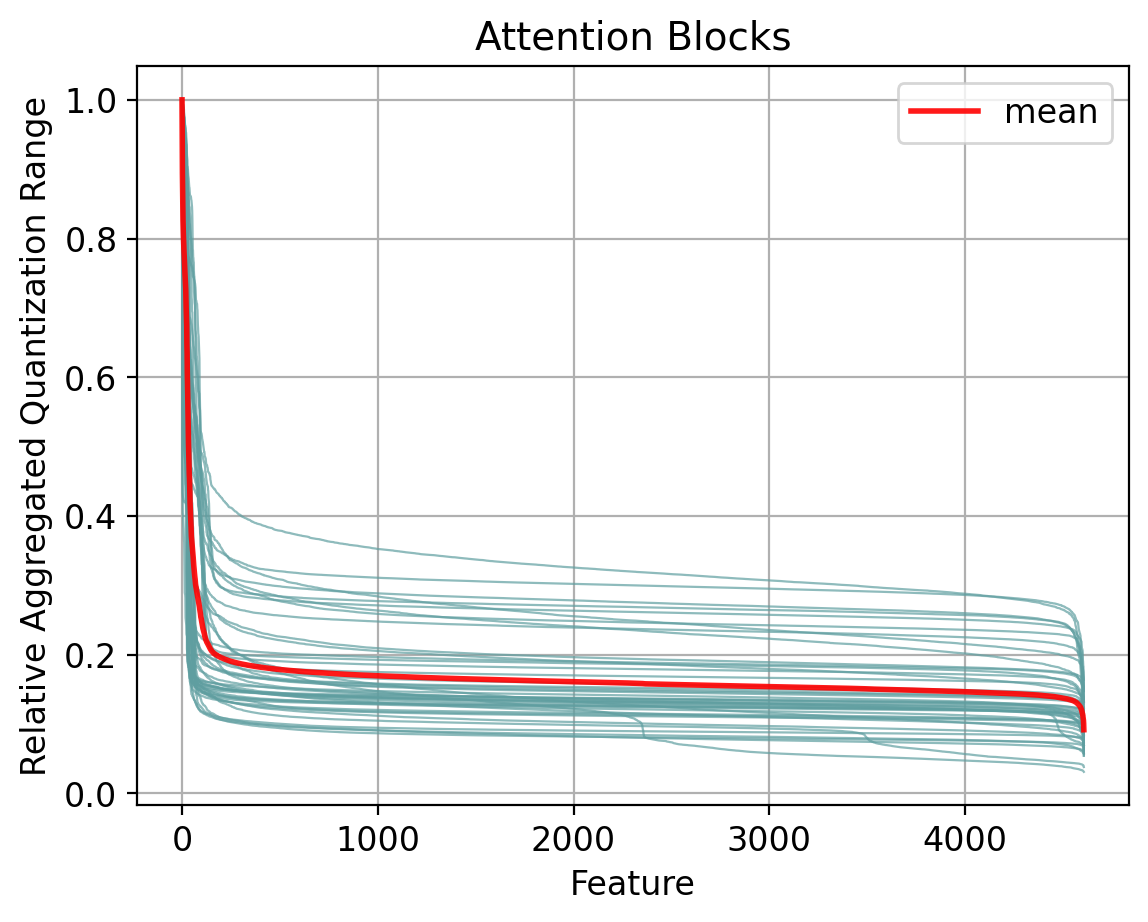}
\hspace{0.05\linewidth}
\includegraphics[width=0.38\linewidth]{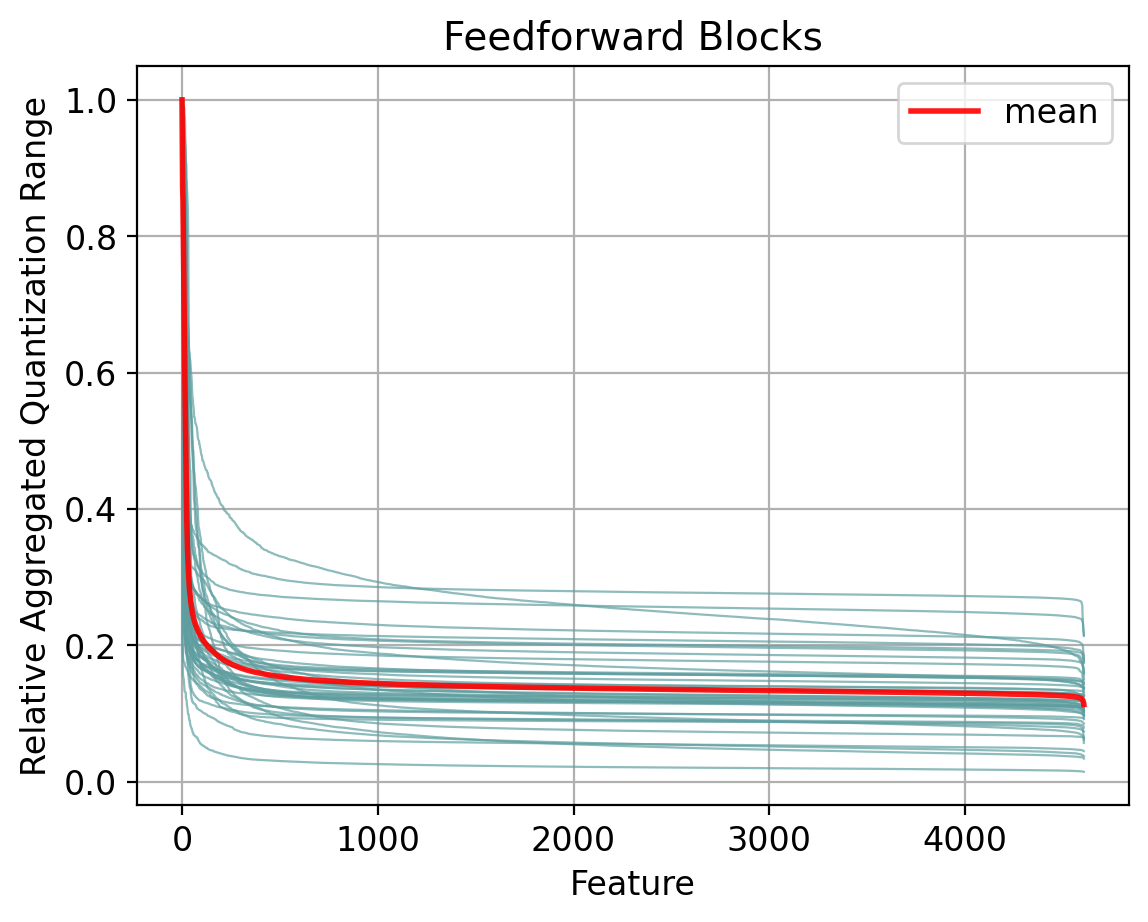}
\vspace{-0.12in}
\caption{Sorted aggregated quantization ranges, $\Bar{\bR}_j$, of each each attention (left) and feedforward (right) block in Gemma 2 27B, with the mean across all layers in red. Values are scaled such that the max range for each layer is set to 1.}
\label{fig:qparam_ranges}
\vspace{-0.12in}
\end{figure}

Thankfully, there are a couple observations that we can leverage. First, the communicated features have consistent structures. Looking at the aggregated quantization ranges for each feature across a calibration set in Figure~\ref{fig:qparam_ranges} (details in Section~\ref{sec:method}), we observe that a small number of features have enormous ranges, potentially resulting in large quantization errors. Second, tensor parallelism can counteract feature quantization error. Theoretically, instead of uniformly distributed quantization errors, quantizing the partial sums prior to synchronization results in aggregated errors that follow the Irwin-Hall distribution, approximately Gaussian as the number of devices increases. \textit{This means tensor parallelism synchronization pushes quantization errors to be clustered around 0.} We take advantage of both observations in our method design.


We propose a solution to reduce tensor parallelism communication cost from 16 bits to less than 4.2 bits per value with minimal performance degradation. Inspired by the extra care taken for outlier features in many LLM quantization methods, the main idea of our algorithm is to determine a static set of features that are kept in BF16 while quantizing everything else to 4 bits without perturbing the weights. In Section~\ref{sec:background}, we formulate the problem with compressed communication for tensor parallelism. Next, Section~\ref{sec:method} illustrates our method to select features to maintain at higher precision and how they are used during inference. Then, Section~\ref{sec:experiments} showcases the performance of our method across multiple LLMs and tasks. For example, our method preserves around 98\% of Gemma 2 27B's original performance despite communicating only about 1/4 of the information.

\section{Tensor Parallelism Synchronization}
\label{sec:background}

Here, we formalize the communication problem in tensor parallelism that we aim to tackle. In tensor parallelism, weights in attention and feedforward blocks are partitioned across devices such that each partition can be computed in parallel until a synchronization step aggregates all outputs together on each device. Attention blocks can be split across the head dimension, and feedforward blocks can be split across the intermediate feature dimension. Synchronization in the form of an AllReduce is necessary after the output projection in attention blocks and after the down projection in feedforward blocks. With this setup, a tensor parallelized model and the original model produce the same outputs.

Define the final linear layer (i.e., the layer immediately before synchronization) in attention blocks or feedforward blocks as $f(\bx) = \bx \bW + \bb$ for $\bx \in \RR^{1 \times D}$, weight $\bW \in \RR^{D \times E}$, and bias $\bb \in \RR^{1 \times E}$ when on a single device. When tensor parallelized, the input and weight take the form $\bx^{(i)} \in \RR^{1 \times \frac{D}{N}}$ and $\bW^{(i)} \in \RR^{\frac{D}{N} \times E}$, respectively, for $N$ devices. The pre-sync point linear layer on each device is:
\begin{align*}
    f^{(i)}(\bx^{(i)}) = \bx^{(i)} \bW^{(i)},
\end{align*}
and a sum of the outputs from all devices produces the same activations as the original layer:
\begin{align*}
    f(\bx) = \left[\sum_{i=1}^N f^{(i)}(\bx^{(i)}) \right] + \bb.
\end{align*}
This operation requires communicating $f^{(i)}(\bx^{(i)}) \in \RR^E$ for each device, which can be expensive for large batch sizes and long sequences. Therefore, we aim to find compression and decompression functions, $\cC^{(i)}$ and $\cD^{(i)}$, so that we can communicate just $\cC^{(i)}(f^{(i)}(\bx^{(i)}))$ between devices and decompress as needed. Good functions should satisfy for all $i$ and $\bx^{(i)}$:
\begin{align*}
    f^{(i)}(\bx^{(i)}) \approx \cD^{(i)}(\cC^{(i)}(f^{(i)}(\bx^{(i)}))).
\end{align*}
We choose $\cC^{(i)}$ and $\cD^{(i)}$ to be quantization and dequantization operations based on the intuition that tensor parallelism can alleviate some of the quantization error in low-bit features.


\section{Method}
\label{sec:method}

\begin{figure}
\centering
\includegraphics[width=0.92\linewidth]{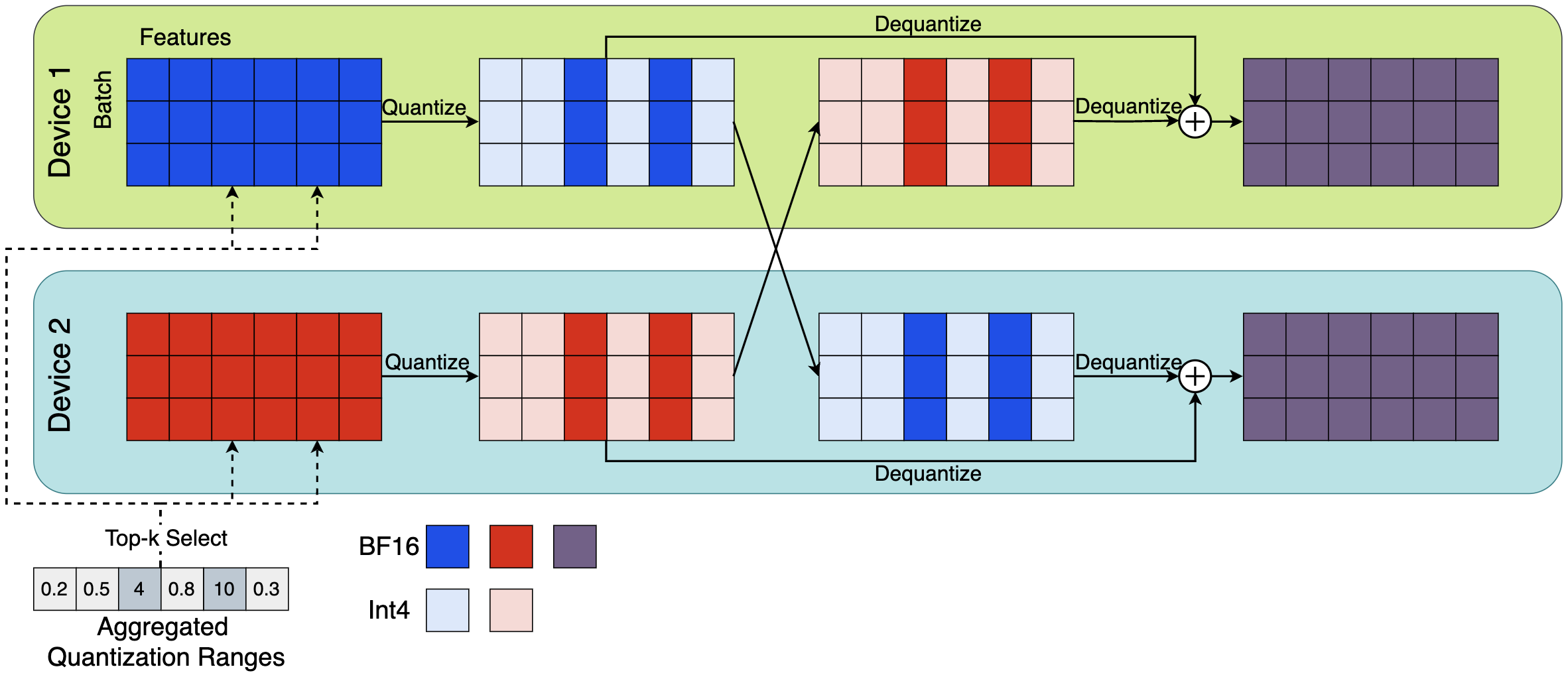}
\vspace{-0.04in}
\caption{Our hybrid quantization algorithm. A small set of features are selected based on aggregated quantization ranges to be kept at BF16 while all others are quantized to Int4 prior to inter-device communication. Then, all tensors are converted to BF16 and summed to sync across each device.}
\label{fig:algorithm}
\vspace{-0.1in}
\end{figure}

Outlined in Figure~\ref{fig:algorithm}, we choose to transfer a fraction of features in BF16 and the rest in 4-bit precision. The BF16 features are chosen based on the range from the calibration set since a larger range would result in greater quantization error. 

\paragraph{Calibration.} We let each feature on each device have static and independent quantization parameters (i.e., $NE$ sets of quantization parameters per tensor parallel block). These parameters are exponential moving averages of the minimum and maximum values that are obtained for each feature on each device on a calibration set. In other words, we find quantization parameters $(\bbm^{(i)}_j, \bM^{(i)}_j)$ for $1 \leq i \leq N$ and $1 \leq j \leq E$ following the recursive update rules:
\begin{align}
    \bbm^{(i)}_j &= (1-\gamma) \bbm^{(i)}_j + \gamma \min(\bY^{(i)}_{\cdot, j}) \label{eq:q_min_update} \\
    \bM^{(i)}_j &= (1-\gamma) \bM^{(i)}_j + \gamma \max(\bY^{(i)}_{\cdot, j}) \label{eq:q_max_update}
\end{align}
for a sequence's features to communicate $\bY^{(i)} = f^{(i)}(\bX^{(i)}) \in \RR^{S \times E}$ and constant $\gamma$. For the first sequence, $\bbm^{(i)}_j = \min(\bY^{(i)}_{\cdot, j})$ and $\bM^{(i)}_j = \max(\bY^{(i)}_{\cdot, j})$. Using symmetric quantization, the range on each device is $\bR^{(i)}_j = 2 \max (-\bbm^{(i)}_j, \bM^{(i)}_j)$. For our experiments, we use 256 random WikiText \citep{merity2016pointer} sequences for calibration following update rules \eqref{eq:q_min_update} and \eqref{eq:q_max_update} with $\gamma = 0.01$.  

\paragraph{Selecting High Precision Features.} Looking at the aggregated quantization ranges across devices after calibration, $\Bar{\bR}_j \coloneqq \sum_{i=1}^N \bR^{(i)}_j$, in Figure~\ref{fig:qparam_ranges}, we see a problem: a small set of features have wide ranges which harms the quantization quality. As a solution, we select the top-$k$ features based on $\Bar{\bR}_j$ to be communicated at higher precision, fixed across all sequences and devices. All other features are symmetrically quantized to Int4. Compared to plain quantization, the additional overhead of our method includes the pre-sync BF16 feature selection and post-sync concatenation of features.

\section{Experiments}
\label{sec:experiments}

Evaluating on multiple LLMs and tasks, we demonstrate that our quantization method preserves nearly all of the original performance at less than 4.2 bits per value. Furthermore, we show that even at lower and higher precision quantization, we still outperform all our baselines. 

Our two baselines and our method use the same symmetric quantization parameters per model. For the first baseline, we quantize everything to Int4. For the second baseline, we randomly select $k$ features with uniform probability to be kept at BF16, and everything else is quantized to Int4. For all experiments, our method and the second baseline fix $k = \lfloor E/64 \rfloor$ which brings the average bits per value to under 4.2 for both. The choice of $1/64$ as the fraction of BF16 features can be substituted, but we found this fraction to strike a good balance between performance and compression (Figure~\ref{fig:qparam_ranges} suggests $k$ can be relatively small). Future work can explore adaptively varying $k$ per layer or input.

\begin{figure}
\vspace{-0.1in}
\centering
\includegraphics[width=0.32\linewidth]{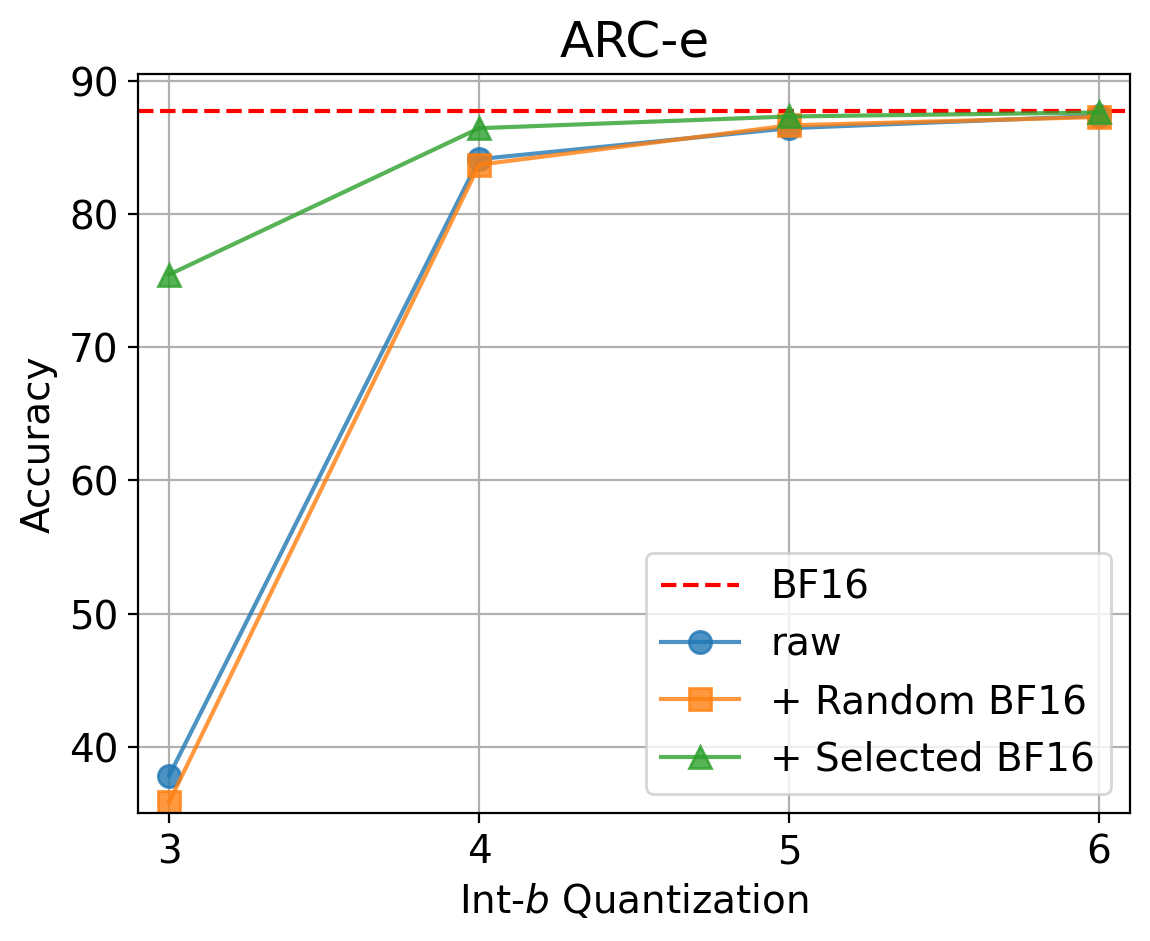}
\includegraphics[width=0.32\linewidth]{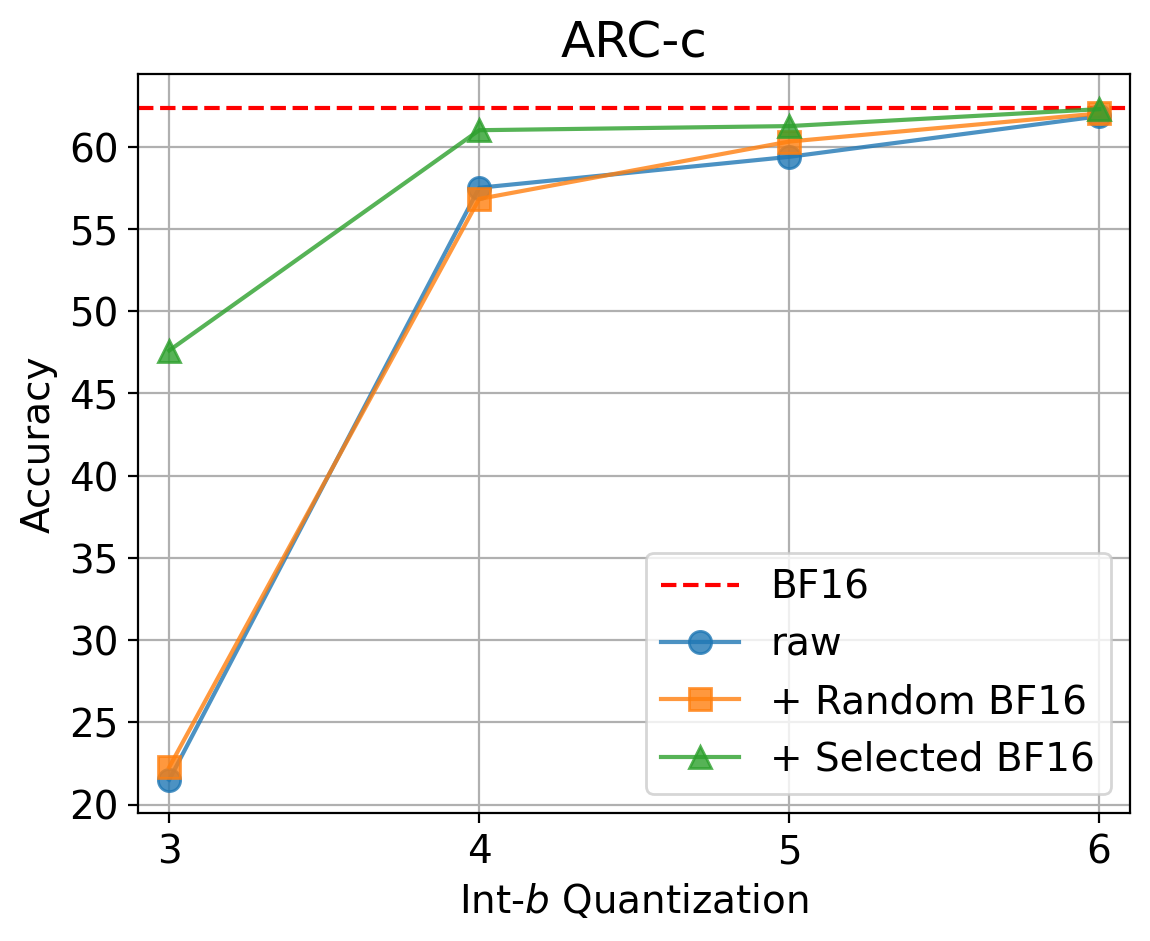}
\includegraphics[width=0.32\linewidth]{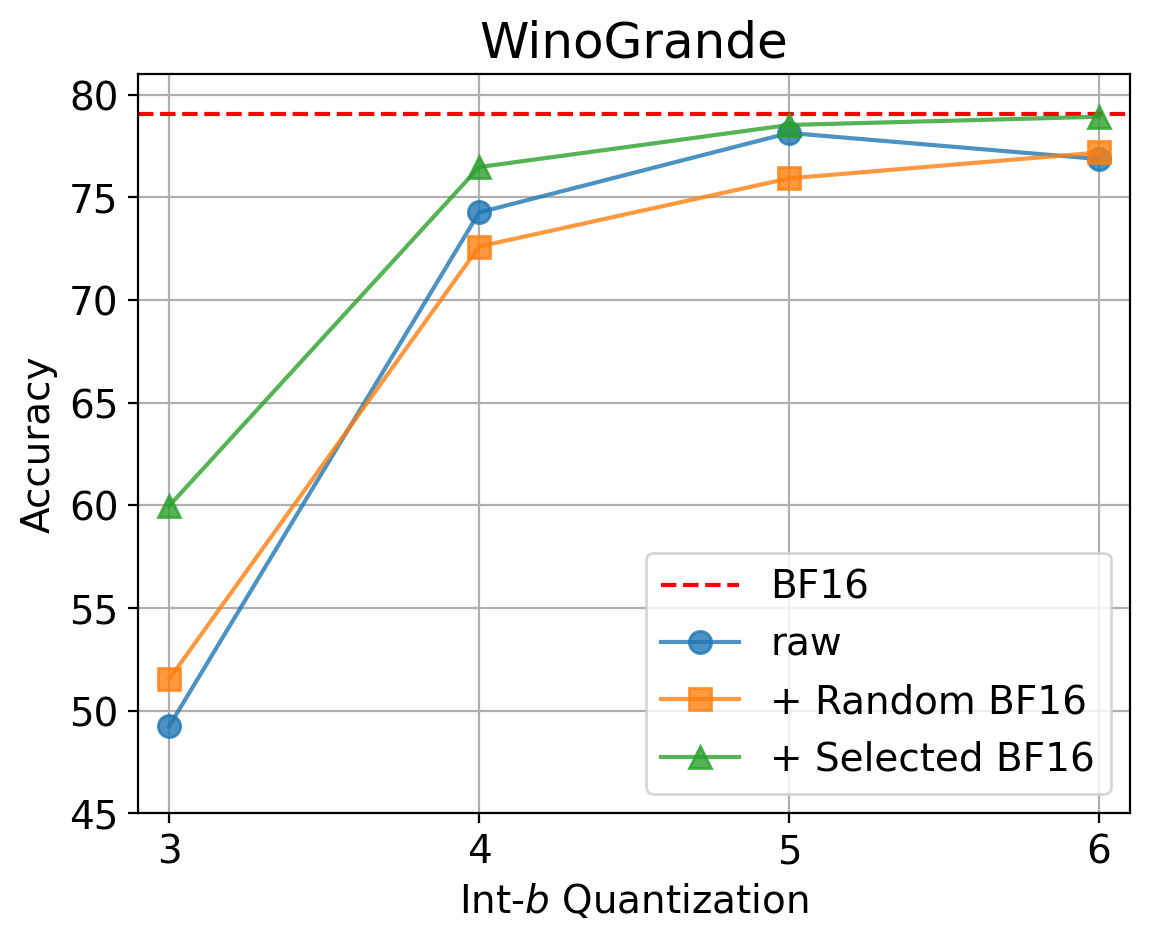}
\caption{Gemma 2 27B performance when features are quantized to varying numbers of bits. Our method achieves the best accuracy for every quantization precision.}
\vspace{-0.1in}
\label{fig:quant_vs_performance}
\end{figure}

Using Gemma 2 27B \citep{team2024gemma}, Llama 2 13B \citep{touvron2023llama}, and Mistral NeMo 12B \citep{mistralnemo}, we compare our method against the baselines and full models on ARC-easy/challenge \citep{allenai:arc}, WinoGrande \citep{sakaguchi2021winogrande}, HellaSwag \citep{zellers2019hellaswag}, and BoolQ \citep{clark2019boolq}. Experimental results are reported for tensor parallelism across 8 devices. From Table~\ref{tab:performance}, we see that our method achieves the best performance for the vast majority of tasks and models among the quantization schemes. Moreover, performance degradation with our method is fairly small in most cases. Overall, our method preserves around 98.0\%, 99.5\%, and 97.1\% of the original Gemma 2 27B, Llama 2 13B, and Mistral NeMo 12B performance in Table~\ref{tab:performance}, respectively. We also observe that choosing random features to be sent in BF16 adds virtually no benefit on top of pure Int4 quantization while choosing features based on aggregated quantization ranges adds a clear performance boost. In fact, in some cases, random selection appears to degrade performance in comparison to pure Int4 quantization, suggesting that maintaining low magnitude features at high precision without preserving high magnitude ones to be harmful. Beyond Int4 quantization, our method consistently best preserves performance at lower and higher precision, as seen in Figure~\ref{fig:quant_vs_performance}.

\begin{table}
\caption{Zero-shot accuracy. The best values for each model and task, excluding the performance of full communication, are in bold. We fix $k = \lfloor E/64 \rfloor$ for Random BF16 and Selected BF16 (ours).}
\label{tab:performance}
\centering
\begin{tabular}{lcccccccc}
\toprule
Method & Bits & ARC-e & ARC-c & WinoGrande & HellaSwag & BoolQ &  \\

\midrule

Gemma 2 27B             & 16 & 87.71 & 62.37 & 79.08 & 65.38 & 84.67 \\
Int4                    & 4  & 84.13 & 57.51 & 74.27 & 63.50 &  83.55     \\
Int4 + Random BF16    & 4.2 & 83.71 & 56.83 & 72.61 & 63.58 &  82.14 \\
Int4 + Selected BF16           & 4.2 & \textbf{86.45} & \textbf{61.01} & \textbf{76.48} &  \textbf{63.86} &  \textbf{83.98} \\

\midrule

Llama 2 13B             & 16 & 79.55 & 48.72 & 72.22 & 60.03 &  80.58 \\
Int4                    & 4  & 77.40 & 47.10 & 68.67 & 58.88 &  78.01 \\
Int4 + Random BF16    & 4.2 & 76.73 & 46.08 & 67.80 & 58.95 &  77.55 \\
Int4 + Selected BF16           & 4.2 & \textbf{79.08} & \textbf{47.35} & \textbf{72.93} & \textbf{59.49} & \textbf{81.22} \\

\midrule

Mistral NeMo 12B       & 16 & 82.91 & 55.80 & 73.01 & 62.82 &  85.17 \\
Int4                    & 4  & 79.50 & 50.77 & 71.43 & 60.97 &  82.84 \\
Int4 + Random BF16    & 4.2 & 79.12 & 51.02 & 70.56 & 60.96 &  81.04 \\
Int4 + Selected BF16           & 4.2 & \textbf{81.06} & \textbf{52.05} & \textbf{71.98} & \textbf{61.74} & \textbf{83.06} \\

\bottomrule
\end{tabular}
\end{table}







\section{Conclusion \& Future Work}
\label{sec:conclusion}

We introduced a method to quantize features for compressed synchronization in tensor parallel LLMs with very little performance degradation. Directly inspired by the consistent nature of outliers in these communicated features, our method combines Int4 and BF16 representations to compress these features to less than 4.2 bits per value. Nevertheless, there are many possibilities for future work. First, we would like to develop a system level implementation of our method to better assess the efficiency gains. Second, our method is fit for AllReduce executed as an AllGather followed by a local reduction, so it would be interesting to see how we can adapt our method to other AllReduce algorithms (e.g. ring-AllReduce). Together, our work and these directions would greatly improve server LLM inference efficiency.

\bibliographystyle{plainnat}
\bibliography{refs}

\end{document}